\documentclass{article}
\usepackage{spconf}
\usepackage{graphicx}
\usepackage{amsmath}
\usepackage{amsfonts}
\usepackage{amssymb}
\usepackage[hidelinks]{hyperref}
\usepackage{subfigure}
\usepackage{mathtools}
\usepackage{algorithm}
\usepackage{algorithmic}
\usepackage{booktabs}
\usepackage{setspace}

\def\x{{\mathbf x}}
\def\z{{\mathbf z}}

\title{CAE-ADMM: Implicit Bitrate Optimization via ADMM-based pruning in Compressive Autoencoders}
\twoauthors
  {Haimeng Zhao}
	{Shanghai High School\\
	Shanghai, China}
  {Peiyuan Liao}
	{Kent School\sthanks{Work done as a senior researcher at Kent Artificial Intelligence Laboratory.}\\
	Kent, CT, USA}
\begin{document}
%
\maketitle
\begin{abstract}
We introduce ADMM-pruned Compressive AutoEncoder (CAE-ADMM) that uses Alternative Direction Method of Multipliers (ADMM) to optimize the trade-off between distortion and efficiency of lossy image compression. Specifically, ADMM in our method is to promote sparsity to implicitly optimize the bitrate, different from entropy estimators used in the previous research. The experiments on public datasets show that our method outperforms the original CAE and some traditional codecs in terms of SSIM/MS-SSIM metrics, at reasonable inference speed. 
\end{abstract}

\section{Introduction}
\label{sec:intro}
Since the proposal of compressive autoencoder (CAE) by \cite{theis2017lossy}, deep learning-approaches have been largely successful in the field of lossy image compression, where its adaptive feature learning capabilities have helped it outperform traditional codecs such as JPEG \cite{wallace1992jpeg} and JPEG 2000 \cite{Taubman2002JPEG}. The goal of such network is to optimize the trade-off between the amount of distortion and the efficiency of the compression, usually expressed by the bitrate or bits per pixel (bpp). In other words, we aim to minimize
\begin{equation}
     d(\mathbf{x},\hat{\mathbf{x}})+\beta \cdot R(\mathbf{\hat{\z}}),
\end{equation}
where $d$ measures the distortion between the input image $\mathbf{x}$ and the reconstructed $\hat{\x}$, $\beta>0$ controls the proportion, and $R$ measures the bitrate of the quantized latent code $\hat{\z}$.

However, this objective is inherently non-differentiable due to the discrete nature of bitrate and quantization. So, to make the problem well-defined so that back-propagation of neural networks is applicable, one of the prevalent solutions proposed by \cite{theis2017lossy} includes a combination of entropy coding and entropy rate estimation, in which they trained a parameterized entropy estimator $H: \mathbb{Z}^M \rightarrow [0,1]$ along with the encoder/decoder that is further made differentiable by an upper-bounding process and the usage of Gaussian scale mixtures (GSMs). 

\begin{figure}[htbp]
    \centering
    \includegraphics[width=\linewidth]{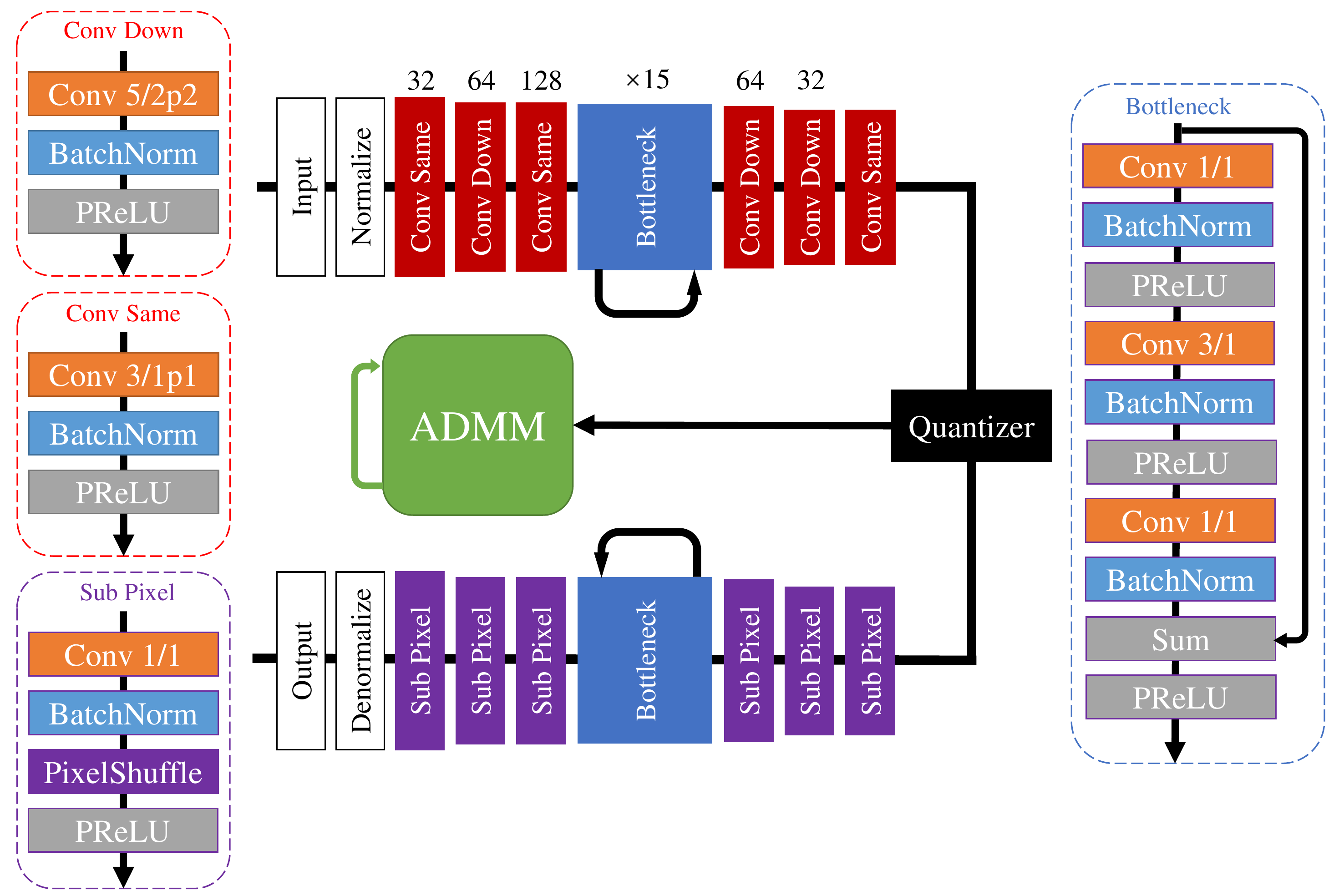}
    \caption{The architecture of CAE-ADMM. ``Conv k/spP'' stands for a convolutional layer with kernel size k$\times$k with a stride of s and a reflection padding of P, and ``Conv Down'' is reducing the height and weight by 2.}
    \label{fig:model}
\end{figure}

As an alternative to methods mentioned above, we replace the entropy estimator by an alternating direction method of multipliers module, where the aggressive pruning on the latent code encourages sparsity, and therefore aid in the optimization of the bitrate. Our experiments show that this pruning paradigm itself is capable of implicitly optimizing the entropy rate while yielding a better result compared with the original CAE and other traditional codecs when measured in both SSIM and MS-SSIM.



\section{Related Works}
\label{sec:relwork}
In the literature, there exists numerous works on variants of compressive autoencoders (CAE) to achieve lossy image compression \cite{theis2017lossy,jiang2017end,balle2016end,agustsson2017soft,li2017learning}, with different approaches to measure the distortion and bitrate. For distance function $ d(\mathbf{x},\hat{\mathbf{x}})$, MSE (Mean-Squared Error), $\text{PSNR} = 10 \cdot \log_{10} (\frac{\text{MAX}_I^2}{\text{MSE}})$ (Peak Signal-to-noise Ratio), $\text{SSIM}(\x,\hat{\x})=\frac{(2\mu_{\x}\mu_{\hat{\x}}+(k_1 L)^2 )(2\sigma_{\x \hat{\x}}+ (k_2 L)^2 )}{ (\mu_{\x}^2 + \mu_{\hat{\x}}^2 + (k_1 L)^2)    (\sigma_{\x}^2 + \sigma_{\hat{\x}}^2 + (k_2 L)^2)   }$ (Structural Similarity Index) and MS-SSIM (Multiscale SSIM) are the most widely used metrics. For the entropy estimator $H$, apart from the GSM model mentioned in the former section, recent works also use generative models\cite{balle2016end} and context models\cite{mentzer2018conditional} as potential alternatives. For works of non-autoencoder approaches, there is also an increasing interest in the usage of GAN \cite{rippel2017real}, GDN (Generalized Divisive Normalization) \cite{balle2016end} and RNN \cite{toderici2015variable}.

The traditional convex optimization ADMM algorithm \cite{ADMM} first saw its application in the field of neural architecture search (NAS) by the method proposed by Han et al. \cite{han2015learning}, which is followed by a few improvements in \cite{Yang2017Designing,Guo2016Dynamic,He2017Channel,ye2018progressive}.



\section{Proposed Method}
\label{sec:method}
A typical CAE consists of an enocder $E$, a decoder $D$ and a quantizer $Q$ \cite{theis2017lossy}:
\begin{align*}
    E:\mathbb{R}^n &\to \mathbb{R}^m,\\
    D:\mathbb{R}^m &\to \mathbb{R}^n,\\
    Q:\mathbb{R}^m &\to \mathbb{Z}^m.
\end{align*}

The encoder $E$ maps the original image $\x \in \mathbb{R}^n$ to a latent representation $\z = E(\x).$ The quantizer $Q$ then maps each element of $\z$ to $\mathbb{Z},$ which produces the compressed presentation of the image $\hat{\z} = Q(\z)$. Finally, the decoder $D$ attempts to reconstruct the original image $\hat{\x} = D(\hat{\z})$ from the information in $\hat{\z}.$

We aim to let the reconstructed image $\hat{\x}$ looks as similar as the original $\x$ (minimize the distance function $d$) while reducing the number of bits needed to store the latent code, or minimize the bitrate $R$. The problem can then be rephrased as below, assuming that $Q$ is not parameterized:

\begin{equation}
    \arg\!\min_{E,D} \ d(\mathbf{x}, D\circ Q \circ E(\mathbf{x}) ) + \beta \cdot R \circ Q \circ E(\mathbf{x}) 
\end{equation}


\subsection{Selection of $E$, $D$ and $Q$}
\label{subs:sel}
Similar to \cite{theis2017lossy}, CAE-ADMM uses convolutional layers to be the basis of our encoder and decoder. The decoder mirrors the structure of the encoder to maintain symmetry, except that uses sub-pixel convolutional layers proposed by Shi et al.\cite{Shi2016Real} to perform up-sampling. 



For the quantizer $Q$, we use a simple and computationally efficient one proposed by Theis et al.\cite{theis2017lossy}, inspired by the random binary version developed by Torderici et al.\cite{toderici2015variable}. It is defined as:
\begin{equation}
    Q(t) = \lfloor t \rfloor + \epsilon,\ \epsilon \in \{0,1\},
\end{equation}

in which $\epsilon$ decides whether to output the ground or the ceiling of the input, and the probability of $\epsilon = 1$ satisfies $P(\epsilon=1)=t - \lfloor t \rfloor.$ To make the quantizer differentiable, we define its gradient with that of its expectation:
\begin{equation}
    \frac{\partial}{\partial{t}} Q(t) =
    \frac{\partial}{\partial{t}} \mathbb{E}[Q(t)] =
    \frac{\partial}{\partial{t}} t = 1
\end{equation}


\subsection{Solution to the optimization problem}
\label{subs:opt}

Since multiple well-defined metrics (mentioned in section \ref{sec:relwork}) exist for $d$, we here aim to provide an alternative method to optimize $R$ without the use of $H$.
Intuitively, we can reformulate $R$ by 
\begin{equation}
    R(\hat{\z})=\text{card}(\hat{\z})
    =\text{card}(Q \circ E(\x)),
\end{equation}
in which $\text{card}(\cdot)$ counts the number of non-zero elements. If we want $z$ generated by the encoder to have fewer number of non-zero elements than a desired number $\ell$, we can rephrase the problem into an ADMM-solvable problem \cite{ye2018progressive}:

\begin{equation}
\begin{split}
     \arg\!\min_{E,D} \ d(\mathbf{x}, \mathbf{\hat{x}}) & + g(\mathbf{Z}) ,\\
    \text{s.t.} \ Q \circ E(\textbf{x})- & \mathbf{Z}  =0.
\end{split}
\end{equation}
where the indicator function $g(\cdot)$ is defined as 

\begin{equation}
    g(\mathbf{Z}) =
    \left\{
        \begin{array}{lr}
            0  &\text{if card}(\mathbf{Z})\leq\ell,\\
            +\infty &\text{otherwise}.
        \end{array}
    \right.
\end{equation}
Remark that both $\mathbf{U}$ and $\mathbf{Z}$ are initialized to be all-zero, and $\mathbf{Z}$ is an element of $\textbf{S}=\{\mathbf{Z}\vert~\text{card}(\mathbf{Z})\leq\ell\}$. By introducing the dual variable $\textbf{U}$ and a penalty factor $\rho>0$, we can split the above problem into two sub-problems.
The first sub-problem is: 

\begin{equation}
    \arg\!\min_{E,D} \ d(\textbf{x},\hat{\textbf{x}}) + \frac{\rho}{2}\Vert Q \circ E(\textbf{x})-\textbf{Z}^k+\textbf{U}^k \Vert^2_{F},
\end{equation}
in which $k$ is the current iteration number and $\Vert\cdot\Vert^2_F$ is the Frobenius norm. This is the neural network's loss with $L_2$ regularization, which can be solved by back propagation and gradient descent.
The second sub-problem is:

\begin{equation}
    \arg\!\min_{Z} \ g(\textbf{Z})+\frac{\rho}{2}\Vert Q^{k+1} \circ E^{k+1}(\textbf{x})-\textbf{Z}+\textbf{U}^k \Vert^2_{F}.
\end{equation}

The solution to this problem was derived by Boyd et al. in 2011\cite{Boyd2011Distributed}:

\begin{equation}
    \textbf{Z}^{k+1}=\Pi_\textbf{S}(Q^{k+1} \circ  E^{k+1}(\textbf{x})+\textbf{U}^k),
\end{equation}

where $\Pi_\textbf{S}(\cdot)$ represents the Euclidean projection onto the set $\textbf{S}$. Generally, Euclidean projection onto a non-convex set is difficult, but Boyd et al.\cite{Boyd2011Distributed} have proved that the optimal solution is to keep the $\ell$ largest elements of $Q^{k+1} \circ E^{k+1}(\textbf{x})+\textbf{U}^k$ and set the rest to zero.
Finally, we will update the dual variable $\textbf{U}$ with the following policy:
\begin{equation}
    \textbf{U}^{k+1}=\textbf{U}^k+Q^{k+1} \circ E^{k+1}(\textbf{x})-\textbf{Z}^{k+1}.
\end{equation}

These three steps together form one iteration of the ADMM pruning method. Algorithm \ref{alg:admm} shows the complete steps.

\begin{algorithm}
\caption{Pruning of CAE Based on ADMM}
\label{alg:admm}
\begin{algorithmic}
\REQUIRE~~\\
$\textbf{x}$: A batch of input images;\\
$E,D$: The encoder and decoder;\\
$Q$: The quantizer; \\
$\ell$: The expected number of non-zero elements in $E(\textbf{x})$;\\
$k_m$: Max number of iterations.
\ENSURE~~\\
$E,D$: Trained encoder and decoder;\\
\STATE $\textbf{U},\textbf{Z} \Leftarrow$ zeroes with the same shape as $Q \circ E(\textbf{x})$
\FOR{$1\leq k \leq k_m$}
\STATE $E,D \Leftarrow \arg\!\min_{E,D} \ d(\textbf{x},\hat{\textbf{x}}) + \frac{\rho}{2}\Vert Q \circ E(\textbf{x})-\textbf{Z}+\textbf{U} \Vert^2_{F}$;
\STATE $\textbf{Z} \Leftarrow$ keep the $\ell$ largest elements in $Q \circ E(\textbf{x})+\textbf{U}$ and set the rest to $0$;
\STATE $\textbf{U}\Leftarrow\textbf{U}+Q \circ E(\textbf{x})-\textbf{Z}.$
\ENDFOR
\RETURN $E,D$
\end{algorithmic}
\end{algorithm}



\section{Experiment}
\label{sec:exp}
\subsection{Model architecture}
Our model architecture, shown in Fig. \ref{fig:model}, is a modification of CAE proposed by \cite{theis2017lossy}. The encoder and decoder are composed of convolutional layers as described in Section \ref{subs:sel}. The input image is first down-sampled by three blocks with each containing a convolutional layer, a batch normalization layer and a PReLU layer. Following 15 residual blocks, two more down-sampling convolutional blocks with the last convolutional block are applied, generating $\z$. The quantizer $Q$ then quantizes it and fed into the decoder whose architecture mirrors the encoder.

\subsection{Training}
We use the Adam optimizer \cite{kingma2014adam} with the batch size set to 32 to solve the first sub-problem. Learning rate is set to $4 \cdot 10^{-3}$ and is halved each time the loss has not dropped for ten epochs. Every 20 epochs, the second and third steps of the ADMM pruning method is applied. The distance function used as a part of back-propagation is a linear combination of MSE and differentiable versions of PSNR/SSIM/MS-SSIM, and the training is first warmed up by a scaled MSE alone. The ratio of the number of elements to retain in step two is set to be $10\%$. To enable fine-grained tuning of bpp, we modify the last layer of the encoder. All procedures are implemented in PyTorch and open-sourced\footnote{https://github.com/JasonZHM/CAE-ADMM}. Each model is trained for 300 epochs on 4 NVIDIA GeForce GTX 1080Ti GPUs.

\subsection{Datasets and preprocessing}
We use BSDS500 \cite{martin2001database} as the training set, which contains five hundred $481\times 321$ natural images. The images are randomly cropped to $128\times 128$, horizontally and vertically flipped and then normalized. For the test set, we use the Kodak PhotoCD dataset \footnote{http://r0k.us/graphics/kodak/}, which contains twenty-four $768\times 512$ images.

\begin{figure*}[htbp]
    \centering
    \subfigure{
        \centering
        \includegraphics[width=0.9\linewidth]{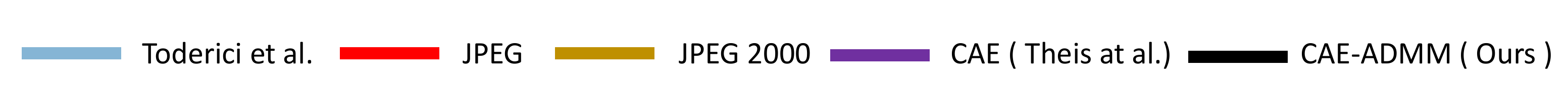}
    }

    \subfigure{
    \begin{minipage}[t]{0.45\linewidth}
        \centering
        \includegraphics[width=\linewidth]{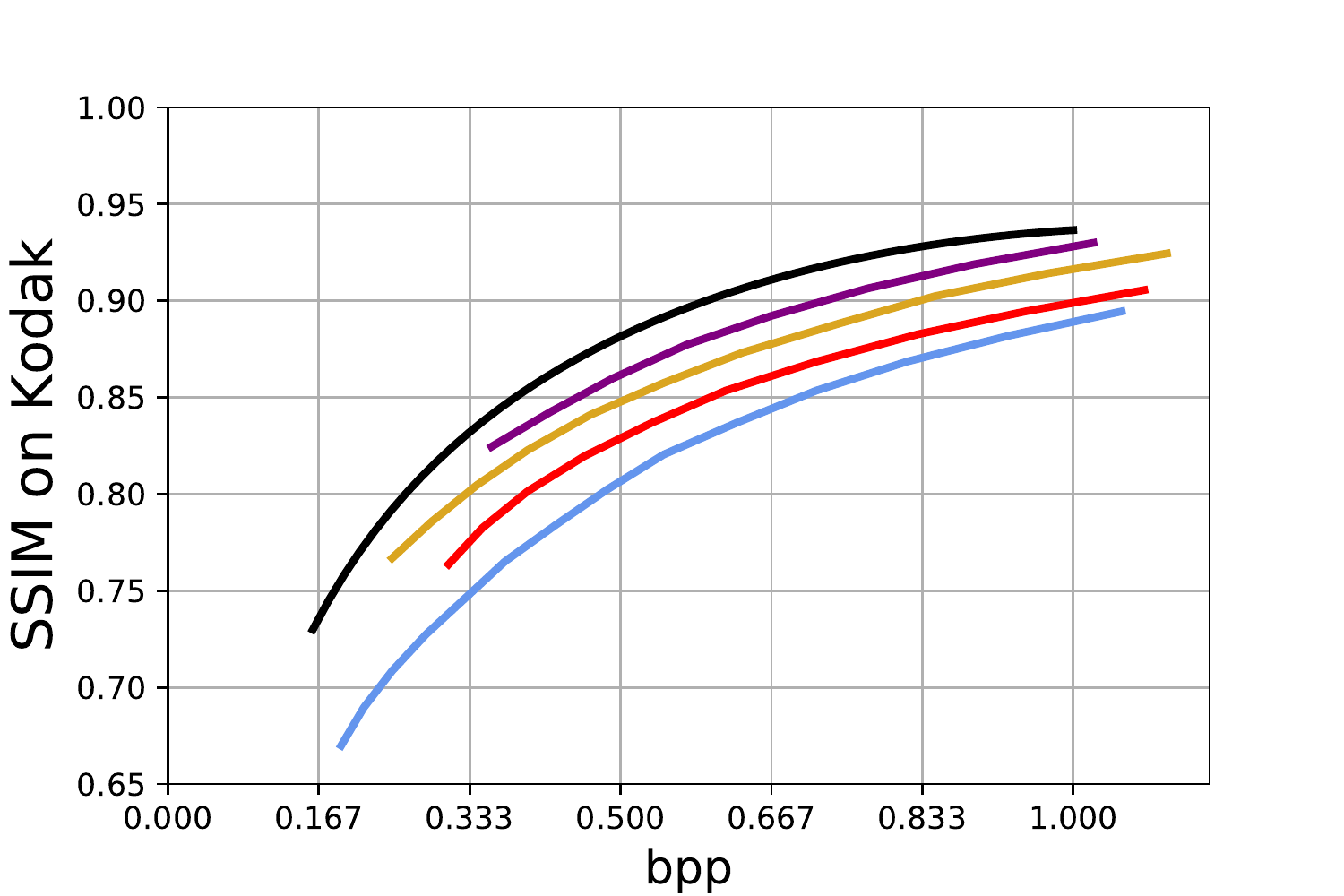}
    \end{minipage}
    \begin{minipage}[t]{0.45\linewidth}
        \centering
        \includegraphics[width=\linewidth]{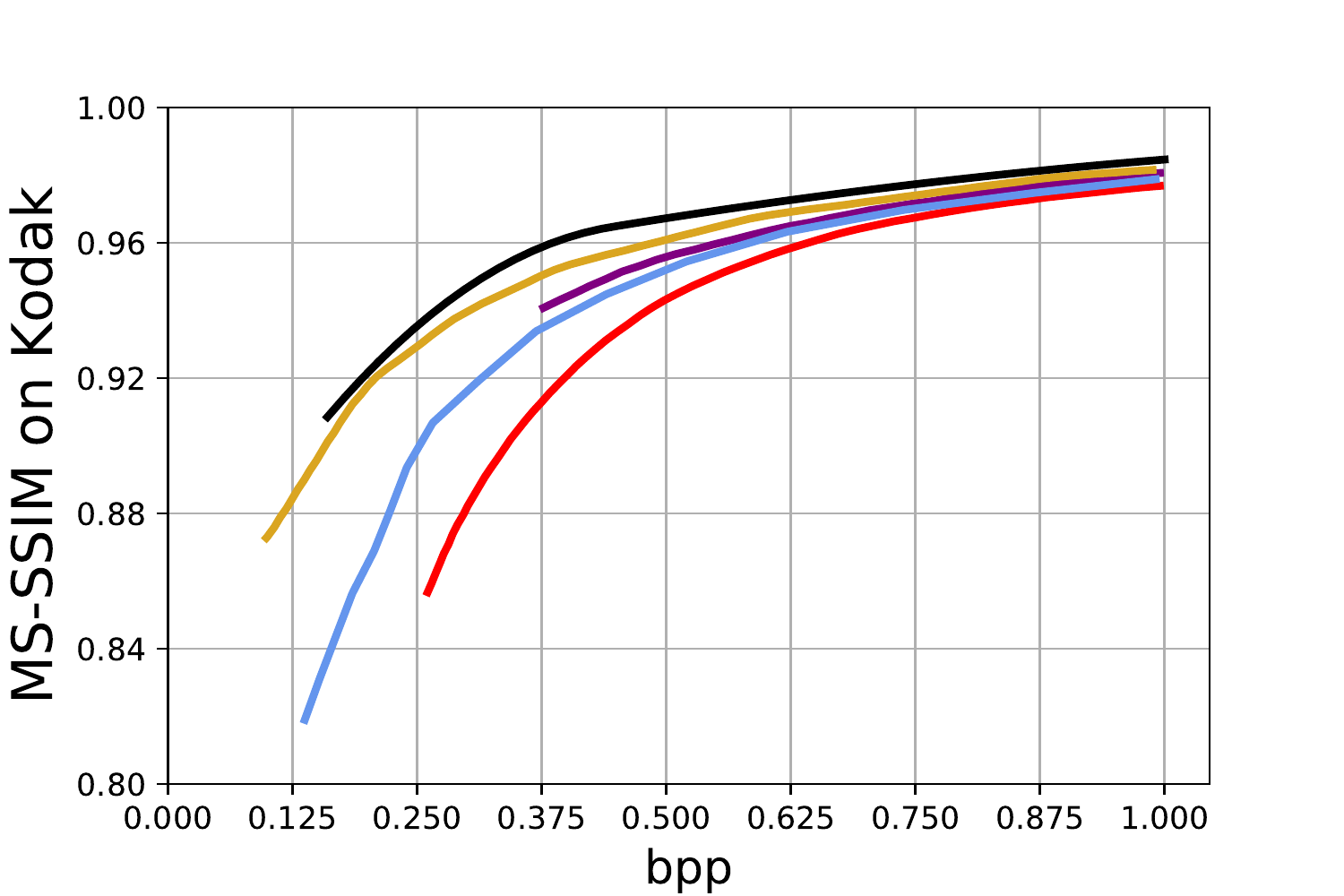}
    \end{minipage}
    }
    \caption{Comparison of different method with respect to SSIM and MS-SSIM on the Kodak PhotoCD dataset. Note that Toderici et al.\cite{toderici2017full} used RNN structure instead of entropy coding while CAE-ADMM (Ours) replaces entropy coding with pruning method.}
    \label{fig:compare}
\end{figure*}

\subsection{Results and discussion}
We test CAE-ADMM (Our method), JPEG (implemented by libjpeg\footnote{http://libjpeg.sourceforge.net/}) and JPEG 2000 (implemented by Kadadu Software\footnote{http://kakadusoftware.com/}) on the Kodak PhotoCD dataset. For the distance metric, we use the open-source implementation of SSIM and MS-SSIM \footnote{https://github.com/jorge-pessoa/pytorch-msssim}.


\begin{figure}[htbp]
    \centering
    \includegraphics[width=\linewidth]{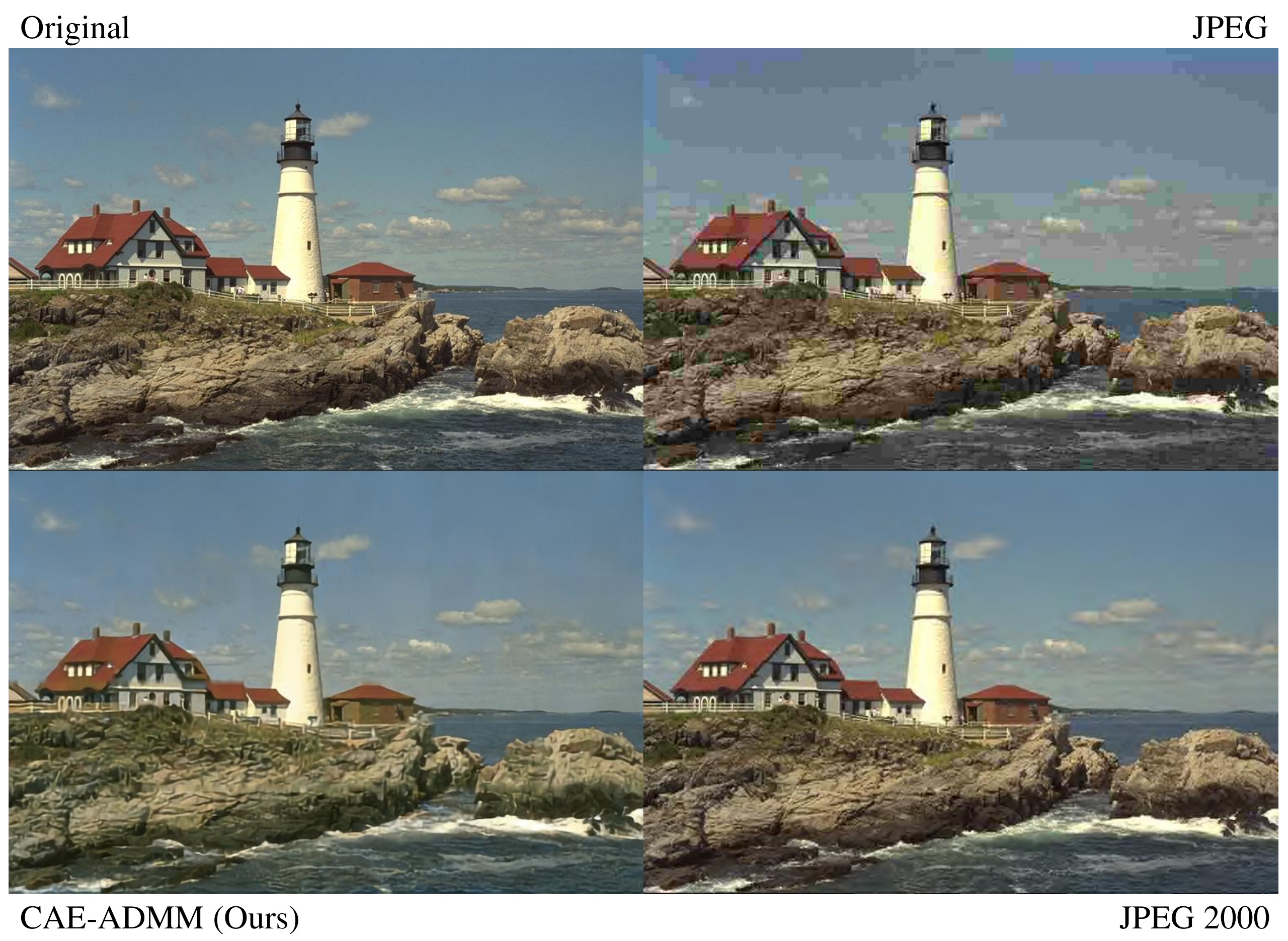}
    \caption{Performance of different methods on image \textit{kodim21} from Kodak dataset. Bpp is set to be about 0.3.}
    \label{fig:cat}
\end{figure}

\begin{figure}[htbp]
    \centering
    \includegraphics[width=\linewidth]{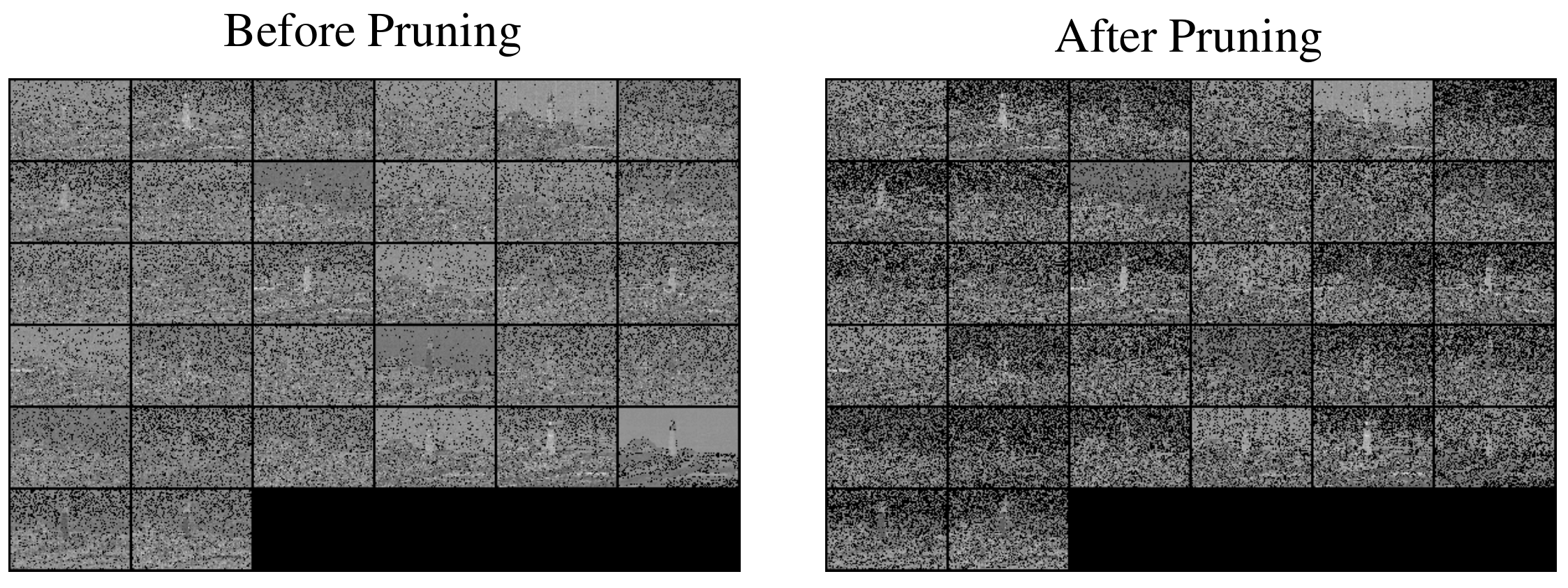}
    \caption{Comparison of latent code before and after pruning for \textit{kodim21}. For the sake of clarity, we marked zero values in the feature map before normalization as black.}
    \label{fig:latent}
\end{figure}

Fig. \ref{fig:compare} shows a comparison of the performance achieved by the mentioned methods on Kodak. Our method (CAE-ADMM) outperforms all the other methods in both SSIM and MS-SSIM, especially the original CAE which uses entropy coding. Note that the blue curve represents the RNN-based method proposed by Toderici et al. which is optimized without an entropy estimator. 

In Fig. \ref{fig:cat}, we demonstrate the effect of different compression methods visually: the origin (top left), JPEG (top right), CAE-ADMM (ours, bottom left) and JPEG 2000 (bottom right). From the figure, we can see that JPEG breaks down under a bpp of 0.3 while that of CAE-ADMM and JPEG 2000 are still satisfactory. 


\begin{table}[htbp]
    \centering
    \begin{tabular}{lcc}
        \toprule
        Model &bpp &ratio of zeros\\
        \midrule
        Before pruning & $1.684 \pm 0.012$ & $7.80\% \pm 3.44\%$ \\
        After pruning  & $1.257 \pm 0.011$ & $17.65\% \pm 4.90$\% \\
        \bottomrule
    \end{tabular}
    \caption{Bpp \& proportion of zero elements in $\hat{\z}$ the total number of elements in $\hat{\z}$ before and after pruning. For both statistics, a 95\% confidence interval is established with a sample size of 233 (size of the mixed dataset). }
    \label{tab:pruning}
\end{table}

For ablation study, we test out the effectiveness of ADMM-module by applying the same training procedure to the same model, one with the pruning schedule and another without. Then, we calculate the average bpp as well as the ratio of zero elements in a mixed dataset ($768\times512$ crops of images from Urban100 \cite{MSLapSRN}, Manga109 \cite{MSLapSRN} and Kodak PhotoCD). Results can be seen in Table \ref{tab:pruning} and more direct visualization of a sample image can be found in Figure \ref{fig:latent}.

Inference-speed-wise, from Table \ref{tab:inference} we can see that our CAE-ADMM has an acceptable inference speed comparing to traditional codecs while maintaining superior quality concerning SSIM.

\begin{table}[htbp]
    \centering
    \begin{tabular}{lccc}
        \toprule
        Model & $\overline{\text{bpp}}$ & SSIM & Second/image\\
        \midrule
        bpp\_0.5 &$\mathbf{0.597}$ & $\mathbf{0.871\!\pm\!0.003}$ &$0.140\!\pm\!0.008$\\
        JPEG     &0.603 & $0.828\!\pm\!0.006$ &$\mathbf{0.033\!\pm\!0.001}$\\
        JPEG2000 &0.601 & $0.793\!\pm\!0.013$ &$0.177\!\pm\!0.020$\\
        \bottomrule
    \end{tabular}
    \caption{95\% confidence intervals of SSIM and inference speed of CAE-ADMM model (inference time being the sum of all $128 \times 128$ patches with batch size = 8) on mixed dataset, 2 GPUs compared to traditional codecs.}
    \label{tab:inference}
\end{table}

\section{Conclusion}
\label{sec:disc}
In this paper, we propose the compressive autoencoder with ADMM-based pruning (CAE-ADMM) \cite{us}, which serves as an alternative to the traditionally used entropy estimating technique for deep-learning-based lossy image compression. Tests on multiple datasets show better results than the original CAE model along with other contemporary approaches. We further explore the effectiveness of the ADMM-based pruning method by looking into the detail of learned latent codes.

Further study can focus on developing a more efficient pruning method, e.g., introducing ideas from the field of reinforcement learning. Also, the structures of $E$, $D$, and $Q$ can be further optimized for speed and accuracy.




\vfill
\bibliographystyle{IEEEbib}
\bibliography{ref}

\end{document}